# Methods to integrate a language model with semantic information for a word prediction component

**Tonio Wandmacher**
Laboratoire d'Informatique (LI)
Université François Rabelais de Tours
3 place Jean-Jaurès, 41000 Blois, France
`tonio.wandmacher@`
`univ-tours.fr`

**Jean-Yves Antoine**
Laboratoire d'Informatique (LI)
Université François Rabelais de Tours
3 place Jean-Jaurès, 41000 Blois, France
`jean-yves.antoine@`
`univ-tours.fr`

## Abstract

Most current word prediction systems make use of n-gram language models (LM) to estimate the probability of the following word in a phrase. In the past years there have been many attempts to enrich such language models with further syntactic or semantic information. We want to explore the predictive powers of Latent Semantic Analysis (LSA), a method that has been shown to provide reliable information on long-distance semantic dependencies between words in a context. We present and evaluate here several methods that integrate LSA-based information with a standard language model: a *semantic cache*, *partial reranking*, and different forms of interpolation. We found that all methods show significant improvements, compared to the 4-gram baseline, and most of them to a simple cache model as well.

## 1   Introduction: NLP for AAC systems

Augmented and Alternative Communication (AAC) is a field of research which concerns natural language processing as well as human-machine interaction, and which aims at restoring the communicative abilities of disabled people with severe speech and motion impairments. These people can be for instance cerebrally and physically handicapped persons or they suffer from a locked-in syndrome due to a cerebral apoplexy. Whatever the disease or impairment considered, oral communication is impossible for these persons who have in addition serious difficulties to control physically their environment. In particular, they are not able to use standard input devices of a computer. Most of the time, they can only handle a single switch device. As a result, communicating with an AAC system consists of typing messages by means of a virtual table of symbols (words, letters or icons) where the user successively selects the desired items.

Basically, an AAC system, such as *FASTY* (Trost et al. 2005) or S<small>IBYLLE</small> (Schadle et al, 2004), consists of four components. At first, one finds a physical input interface connected to the computer. This device is adapted to the motion capacities of the user. When the latter must be restricted to a single switch (eye glimpse or breath detector, for instance), the control of the environment is reduced to a mere *Yes/No* command.

Secondly, a virtual keyboard is displayed on screen. It allows the user to select successively the symbols that compose the intended message. In S<small>IBYLLE</small>, key selection is achieved by pointing letters through a linear scan procedure: a cursor successively highlights each key of the keyboard.

The last two components are a text editor (to write e-mails or other documents) and a speech synthesis module, which is used in case of spoken communication. The latest version of S<small>IBYLLE</small> works for French and German, and it is usable with any *Windows*™ application (text editor, web browser, mailer...), which means that the use of a specific editor is no longer necessary.

The main weakness of AAC systems results from the slowness of message composition. On average, disabled people cannot type more than 1 to 5 words per minute; moreover, this task is very tiring. The use of NLP techniques to improve AAC systems is therefore of first importance.

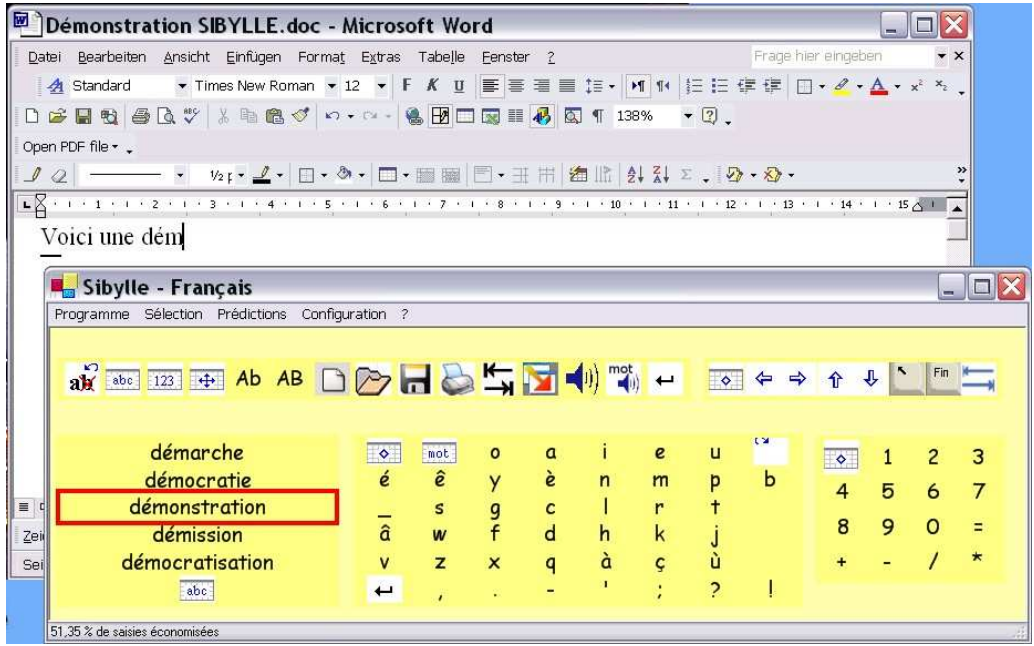

Figure 1: User interface of the *SIBYLLE* AAC system

Two complementary approaches are possible to speed up communication. The first one aims at minimizing the duration of each item selection. Considering a linear scan procedure, one could for instance dynamically reorganize the keyboard in order to present the most probable symbols at first. The second strategy tries to minimize the number of keystrokes to be made. Here, the system tries to predict the words which are likely to occur just after those already typed. The predicted word is then either directly displayed after the end of the inserted text (a method referred to as "word completion", cf. Boissière and Dours, 1996), or a list of N-best (typically 3 to 7) predictions is provided on the virtual keyboard. When one of these predictions corresponds to the intended word, it can be selected by the user. As can be seen in figure 1, the interface of the *SIBYLLE* system presents such a list of most probable words to the user.

Several approaches can be used to carry out word prediction. Most of the commercial AAC systems make only use of a simple lexicon: in this approach, the context is not considered.

On the other hand, stochastic language models can provide a list of word suggestions, depending on the *n-1* (typically *n* = 3 or 4) last inserted words. It is obvious that such a model cannot take into account long-distance dependencies. There have been attempts to integrate part-of-speech information (Fazly and Hirst, 2003) or more complex syntactic models (Schadle et al, 2004) to achieve a better prediction. In this paper, we will nevertheless limit our study to a standard 4-gram model as a baseline to make our results comparable. Our main aim is here to investigate the use of long-distance semantic dependencies to dynamically adapt the prediction to the current semantic context of communication. Similar work has been done by Li and Hirst (2005) and Matiasek and Baroni (2003), who exploit *Pointwise Mutual Information* (PMI; Church and Hanks, 1989). Trnka et al. (2005) dynamically interpolate a high number of topic-oriented models in order to adapt their predictions to the current topic of the text or conversation.

Classically, word predictors are evaluated by an objective metric called *Keystroke Saving Rate* (*ksr*):

$$ksr_n = \left(1 - \frac{k_p}{k_a}\right) \cdot 100 \qquad (1)$$

with $k_p$, $k_a$ being the number of keystrokes needed on the input device when typing a message with ($k_p$) and without prediction ($k_a$ = number of characters in the text that has been entered, $n$ = length of the prediction list, usually $n$ = 5). As

Trost et al. (2005) and Trnka et al. (2005), we assume that one additional keystroke is required for the selection of a word from the list and that a space is automatically inserted afterwards. Note also that words, which have already occurred in the list, will not reappear after the next character has been inserted.

The perplexity measure, which is frequently used to assess statistical language models, proved to be less accurate in this context. We still present perplexities as well in order to provide comparative results.

## 2 Language modeling and semantics

### 2.1 Statistical Language Models

For about 10 to 15 years statistical language modeling has had a remarkable success in various NLP domains, for instance in speech recognition, machine translation, Part-of-Speech tagging, but also in word prediction systems. N-gram based language models (LM) estimate the probability of occurrence for a word, given a string of *n*-1 preceding words. However, computers have only recently become powerful enough to estimate probabilities on a reasonable amount of training data. Moreover, the larger *n* gets, the more important the problem of combinatorial explosion for the probability estimation becomes. A reasonable trade-off between performance and number of estimated events seems therefore to be an *n* of 3 to 5, including sophisticated techniques in order to estimate the probability of unseen events (smoothing methods).

Whereas n-gram-like language models are already performing rather well in many applications, their capacities are also very limited in that they cannot exploit any deeper linguistic structure. Long-distance syntactic relationships are neglected as well as semantic or thematic constraints.

In the past 15 years many attempts have been made to enrich language models with more complex syntactic and semantic models, with varying success (cf. (Rosenfeld, 1996), (Goodman, 2002) or in a word prediction task: (Fazly and Hirst, 2003), (Schadle, 2004), (Li and Hirst, 2005)). We want to explore here an approach based on *Latent Semantic Analysis* (Deerwester et al, 1990).

### 2.2 Latent Semantic Analysis

Several works have suggested the use of *Latent Semantic Analysis* (LSA) in order to integrate semantic similarity to a language model (cf. Bellegarda, 1997; Coccaro and Jurafsky, 1998). LSA models semantic similarity based on co-occurrence distributions of words, and it has shown to be helpful in a variety of NLP tasks, but also in the domain of cognitive modeling (Landauer et al, 1997).

LSA is able to relate coherent contexts to specific content words, and it is good at predicting the occurrence of a content word in the presence of other thematically related terms. However, since it does not take word order into account ("bag-of-words" model) it is very poor at predicting their actual position within the sentence, and it is completely useless for the prediction of function words. Therefore, some attempts have been made to integrate the information coming from an LSA-based model with standard language models of the n-gram type.

In the LSA model (Deerwester et al, 1990) a word $w_i$ is represented as a high-dimensional vector, derived by *Singular Value Decomposition* (SVD) from a term × document (or a term × term) co-occurrence matrix of a training corpus. In this framework, a context or history $h$ (= $w_1$, ... , $w_m$) can be represented by the sum of the (already normalized) vectors corresponding to the words it contains (Landauer et al. 1997):

$$\vec{h} = \sum_{i=1}^{m} \vec{w}_i \qquad (2)$$

This vector reflects the meaning of the preceding (already typed) section, and it has the same dimensionality as the term vectors. It can thus be compared to the term vectors by well-known similarity measures (scalar product, cosine).

### 2.3 Transforming LSA similarities into probabilities

We make the assumption that an utterance or a text to be entered is usually semantically cohesive. We then expect all word vectors to be close to the current context vector, whose corresponding words belong to the semantic field of the context. This forms the basis for a simple probabilistic model of LSA: After calculating the cosine similarity for each word vector $\vec{w}_i$ with the vector $\vec{h}$ of the current context, we could use the normalized similarities as probability values. This probability distribution however is usually rather flat (i.e. the dynamic range is low). For this reason a contrasting (or temperature) factor γ is normally applied (cf. Coccaro and Jurafsky, 1998), which raises the cosine to some power (γ is normally between 3 and 8). After normalization we obtain a probability distribution which can be used for prediction purposes. It is calculated as follows:

$$P_{LSA}(w_i|h) = \frac{\left(\cos(\vec{w}_i, \vec{h}) - \cos_{min}(\vec{h})\right)^\gamma}{\sum_k \left(\cos(\vec{w}_k, \vec{h}) - \cos_{min}(\vec{h})\right)^\gamma} \quad (3)$$

$w_i$ is a word in the vocabulary, h is the current context (history) $\vec{w}_i$ and $\vec{h}$ are their corresponding vectors in the LSA space; $\cos_{min}(\vec{h})$ returns the lowest cosine value measured for $\vec{h}$). The denominator then normalizes each similarity value to ensure that $\sum_k^n P_{LSA}(w_k, h) = 1$.

Let us illustrate the capacities of this model by giving a short example from the French version of our own LSA predictor:

Context: "*Mon père était professeur en mathématiques et je pense que* "
("My dad has been a professor in mathematics and I think that ")

| Rank | Word | P |
|---|---|---|
| 1. | *professeur* ('professor') | 0.0117 |
| 2. | *mathématiques* ("mathematics") | 0.0109 |
| 3. | *enseigné* (participle of 'taught') | 0.0083 |
| 4. | *enseignait* ('taught') | 0.0053 |
| 5. | *mathématicien* ('mathematician') | 0.0049 |
| 6. | *père* ('father') | 0.0046 |
| 7. | *mathématique* ('mathematics') | 0.0045 |
| 8. | *grand-père* ('grand-father') | 0.0043 |
| 9. | *sciences* ('sciences') | 0.0036 |
| 10. | *enseignant* ('teacher') | 0.0032 |

Example 1: Most probable words returned by the LSA model for the given context.

As can be seen in example 1, all ten predicted words are semantically related to the context, they should therefore be given a high probability of occurrence. However, this example also shows the drawbacks of the LSA model: it totally neglects the presence of function words as well as the syntactic structure of the current phrase. We therefore need to find an appropriate way to integrate the information coming from a standard n-gram model and the LSA approach.

## 2.4 Density as a confidence measure

Measuring relation quality in an LSA space, Wandmacher (2005) pointed out that the reliability of LSA relations varies strongly between terms. He also showed that the entropy of a term does not correlate with relation quality (i.e. number of semantically related terms in an LSA-generated term cluster), but he found a medium correlation (*Pearson* coeff. = 0.56) between the number of semantically related terms and the average cosine similarity of the *m* nearest neighbors (density). The closer the nearest neighbors of a term vector are, the more probable it is to find semantically related terms for the given word. In turn, terms having a high density are more likely to be semantically related to a given context (i.e. their specificity is higher).

We define the density of a term $w_i$ as follows:

$$D_m(w_i) = \frac{1}{m} \cdot \sum_{j=1}^{m} \cos(\vec{w}_i, NN_j(\vec{w}_i)) \quad (4)$$

In the following we will use this measure (with *m*=100) as a confidence metric to estimate the reliability of a word being predicted by the LSA component, since it showed to give slightly better results in our experiments than the entropy measure.

## 3 Integrating semantic information

In the following we present several different methods to integrate semantic information as it is provided by an LSA model into a standard LM.

### 3.1 Semantic cache model

Cache (or recency promotion) models have shown to bring slight but constant gains in language modeling (Kuhn and De Mori, 1990). The underlying idea is that words that have already occurred in a text are more likely to occur another time. Therefore their probability is raised by a constant or exponentially decaying factor, depending on the position of the element in the cache. The idea of a decaying cache function is that the probability of reoccurrence depends on the cosine similarity of the word in the cache and the word to be predicted. The highest probability of reoccurrence is usually after 15 to 20 words.
Similar to Clarkson and Robinson (1997), we implemented an exponentially decaying cache of length *l* (usually between 100 and 1000), using the following decay function for a word $w_i$ and its position $p$ in the cache.

$$f_d(w_i, p) = e^{\left(\frac{-0.5(p-\mu)}{\sigma}\right)^2} \quad (5)$$

$\sigma = \mu/3$ if $p < \mu$ and $\sigma = l/3$ if $p \geq \mu$. The function returns 0 if $w_i$ is not in the cache, and it is 1 if $p = \mu$. A typical graph for (5) can be seen in figure (2).

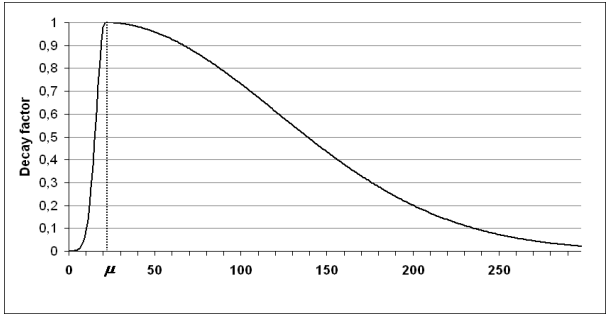

Figure 2: Decay function with $\mu=20$ and $l=300$.

We extend this model by calculating for each element having occurred in the context its $m$ nearest LSA neighbors ($NN_m(\vec{w}_{occ}, \theta)$, using cosine similarity), if their cosine lies above a threshold $\theta$, and add them to the cache as well, right after the word that has occurred in the text ("*Bring your friends*"-strategy). The size of the cache is adapted accordingly (for $\mu$, $\sigma$ and $l$), depending on the number of neighbors added. This results in the following cache function:

$$P_{cache}(w_i) = \sum_{l} \beta \cdot f_{cos}(w^i_{occ}, w_i) \cdot f_d(w_i, p) \quad (6)$$

with $l$ = size of the cache. $\beta$ is a constant controlling the influence of the component (usually $\beta \approx 0.1/l$); $w^i_{occ}$ is a word that has already recently occurred in the context and is therefore added as a standard cache element, whereas $w_i$ is a nearest neighbor to $w^i_{occ}$. $f_{cos}(w^i_{occ}, w_i)$ returns the cosine similarity between $\vec{w}^i_{occ}$ and $\vec{w}_i$, with $cos(\vec{w}^i_{occ}, \vec{w}_i) > \theta$ (Rem: $w_i$ with $cos(\vec{w}^i_{occ}, \vec{w}_i) \leq \theta$ have not been added to the cache). Since $cos(\vec{w}_i, \vec{w}_i)=1$, terms having actually occurred before will be given full weight, whereas all $w_i$ being only nearest LSA neighbors to $w^i_{occ}$ will receive a weight corresponding to their cosine similarity with $w^i_{occ}$, which is less than 1 (but larger than $\theta$).

$f_d(w_i, p)$ is the decay factor for the current position $p$ of $w_i$ in the cache, calculated as shown in equation (5).

### 3.2 Partial reranking

The underlying idea of partial reranking is to regard only the best $n$ candidates from the basic language model for the semantic model in order to prevent the LSA model from making totally implausible (i.e. improbable) predictions. Words being improbable for a given context will be disregarded as well as words that do not occur in the semantic model (e.g. function words), because LSA is not able to give correct estimates for this group of words (here the base probability remains unchanged).

For the best $n$ candidates their semantic probability is calculated and each of these words is assigned an additional value, after a fraction of its base probability has been subtracted (*jackpot* strategy).

For a given context $h$ we calculate the ordered set

$BEST_n(h) = <w_1, \ldots, w_n>$, so that $P(w_1|h) \geq P(w_2|h) \geq \ldots \geq P(w_n|h)$

For each $w_i$ in $BEST_n(h)$ we then calculate its reranking probability as follows:

$$P_{RR}(w_i) = \beta \cdot \cos(\vec{w}_i, \vec{h}) \cdot D(w_i) \cdot I(Best_n(h), w_i) \quad (7)$$

$\beta$ is a weighting constant controlling the overall influence of the reranking process, $cos(\vec{w}_i, \vec{w}_i)$ returns the cosine of the word's vector and the current context vector, $D(w_i)$ gives the confidence measure of $w_i$ and $I$ is an indicator function being 1, iff $w_i \in BEST(h)$, and 0 otherwise.

### 3.3 Standard interpolation

Interpolation is the standard way to integrate information from heterogeneous resources. While for a linear combination we simply add the weighted probabilities of two (or more) models, geometric interpolation multiplies the probabilities, which are weighted by an exponential coefficient ($0 \leq \lambda_1 \leq 1$):

Linear Interpolation (LI):

$$P'(w_i) = \lambda_1 \cdot P_b(w_i) + (1-\lambda_1) \cdot P_s(w_i) \quad (8)$$

Geometric Interpolation (GI):

$$P'(w_i) = \frac{P_b(w_i)^{\lambda_1} \cdot P_s(w_i)^{(1-\lambda_1)}}{\sum_{j=1}^{n} P_b(w_j)^{\lambda_1} \cdot P_s(w_j)^{(1-\lambda_1)}} \quad (9)$$

The main difference between the two methods is that the latter takes the agreement of two models into account. Only if each of the single models assigns a high probability to a given event will the combined probability be assigned a high value. If one of the models assigns a high probability and the other does not the resulting probability will be lower.

### 3.4 Confidence-weighted interpolation

Whereas in standard settings the coefficients are stable for all probabilities, some approaches use confidence-weighted coefficients that are adapted for each probability. In order to integrate n-gram and LSA probabilities, Coccaro and Jurafsky (1998) proposed an entropy-related confidence measure for the LSA component, based on the observation that words that occur in many different contexts (i.e. have a high entropy), cannot well be predicted by LSA. We use here a density-based measure (cf. section 2.2), because we found it more reliable than entropy in preliminary tests. For interpolation purposes we calculate the coefficient of the LSA component as follows:

$$\lambda_i = \beta \cdot D(w_i), \text{ iff } D(w_i) > 0; 0 \text{ otherwise} \quad (10)$$

with $\beta$ being a weighting constant to control the influence of the LSA predictor. For all experiments, we set $\beta$ to 0.4 (i.e. $0 \leq \lambda_i \leq 0.4$), which proved to be optimal in pre-tests.

## 4 Results

We calculated our baseline n-gram model on a 44 million word corpus from the French daily *Le Monde* (1998-1999). Using the *SRI* toolkit (Stolcke, 2002)[1] we computed a 4-gram LM over a controlled 141,000 word vocabulary, using *modified Kneser-Ney* discounting (Goodman, 2001), and we applied *Stolcke* pruning (Stolcke, 1998) to reduce the model to a manageable size ($\theta = 10^{-7}$).

---
[1] SRI Toolkit: www.speech.sri.com.

The LSA space was calculated on a 100 million word corpus from *Le Monde* (1996 – 2002). Using the *Infomap* toolkit[2], we generated a term × term co-occurrence matrix for an 80,000 word vocabulary (matrix size = 80,000 × 3,000), stopwords were excluded. After several pre-tests, we set the size of the co-occurrence window to ±100. The matrix was then reduced by singular value decomposition to 150 columns, so that each word in the vocabulary was represented by a vector of 150 dimensions, which was normalized to speed up similarity calculations (the scalar product of two normalized vectors equals the cosine of their angle).

Our test corpus consisted of 8 sections from the French newspaper *Humanité*, (January 1999, from 5,378 to 8,750 words each), summing up to 58,457 words. We then calculated for each test set the keystroke saving rate based on a 5-word list ($ksr_5$) and perplexity for the following settings[3]:

1. 4-gram LM only (baseline)
2. 4-gram + decaying cache ($l = 400$)
3. 4-gram + LSA using linear interpolation with $\lambda_{LSA} = 0.11$ (LI).
4. 4-gram + LSA using geometric interpolation, with $\lambda_{LSA} = 0.07$ (GI).
5. 4-gram + LSA using linear interpolation and (density-based) confidence weighting (CWLI).
6. 4-gram + LSA using geometric interpolation and (density-based) confidence weighting (CWGI).
7. 4-gram + partial reranking ($n = 1000$, $\beta = 0.001$)
8. 4-gram + decaying semantic cache ($l = 4000$; $m = 10$; $\theta = 0.4$, $\beta = 0.0001$)

Figures 3 and 4 display the overall results in terms of *ksr* and perplexity.

---
[2] Infomap Project: http://infomap-nlp.sourceforge.net/
[3] All parameter settings presented here are based on results of extended empirical pre-tests. We used held-out development data sets that have randomly been chosen from the *Humanité* corpus.(8k to 10k words each). The parameters being presented here were optimal for our test sets. For reasons of simplicity we did not use automatic optimization techniques such as the EM algorithm (cf. Jelinek, 1990).

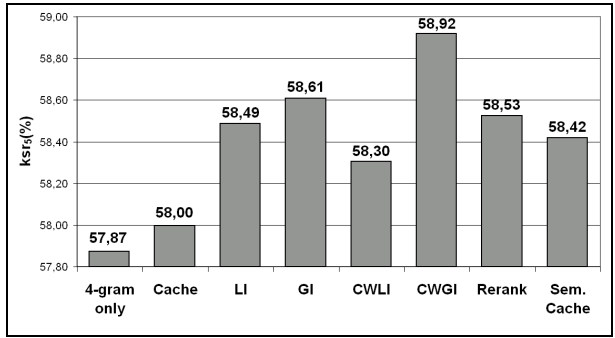

Figure 3: Results ($ksr_5$) for all methods tested.

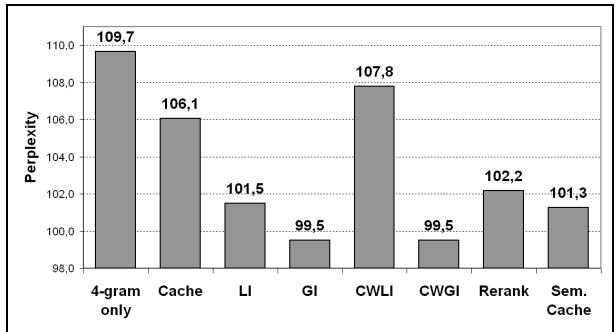

Figure 4: Results (perplexity) for all methods tested.

Using the results of our 8 samples, we performed paired *t* tests for every method with the baseline as well as with the cache model. All gains for *ksr* turned out to be highly significant (sig. level < 0.001), and apart from the results for CWLI, all perplexity reductions were significant as well (sig. level < 0.007), with respect to the cache results. We can therefore conclude that, with exception of CWLI, all methods tested have a beneficial effect, even when compared to a simple cache model. The highest gain in *ksr* (with respect to the baseline) was obtained for the confidence-weighted geometric interpolation method (CWGI; +1.05%), the highest perplexity reduction was measured for GI as well as for CWGI (-9.3% for both). All other methods (apart from IWLI) gave rather similar results (+0.6 to +0.8% in *ksr*, and -6.8% to -7.7% in perplexity).

We also calculated for all samples the correlation between *ksr* and perplexity. We measured a *Pearson* coefficient of -0.683 (Sig. level < 0.0001).

At first glance, these results may not seem overwhelming, but we have to take into account that our *ksr* baseline of 57.9% is already rather high, and at such a level, additional gains become hard to achieve (cf. Lesher et al, 2002).

The fact that CWLI performed worse than even simple LI was not expected, but it can be explained by an inherent property of linear interpolation: If one of the models to be interpolated overestimates the probability for a word, the other cannot compensate for it (even if it gives correct estimates), and the resulting probability will be too high. In our case, this happens when a word receives a high confidence value; its probability will then be overestimated by the LSA component.

## 5 Conclusion and further work

Adapting a statistical language model with semantic information, stemming from a distributional analysis like LSA, has shown to be a non-trivial problem. Considering the task of word prediction in an AAC system, we tested different methods to integrate an n-gram LM with LSA: A semantic cache model, a partial reranking approach, and some variants of interpolation.

We evaluated the methods using two different measures, the keystroke saving rate (*ksr*) and perplexity, and we found significant gains for all methods incorporating LSA information, compared to the baseline. In terms of *ksr* the most successful method was confidence-weighted geometric interpolation (CWGI; +1.05% in *ksr*); for perplexity, the greatest reduction was obtained for standard as well as for confidence-weighted geometric interpolation (-9.3% for both). Partial reranking and the semantic cache gave very similar results, despite their rather different underlying approach.

We could not provide here a comparison with other models that make use of distributional information, like the trigger approach by Rosenfeld (1996), Matiasek and Baroni (2003) or the model presented by Li and Hirst (2005), based on *Pointwise Mutual Information* (PMI). A comparison of these similarities with LSA remains to be done.

Finally, an AAC system has not only the function of simple text entering but also of providing cognitive support to its user, whose communicative abilities might be totally depending on it. Therefore, she or he might feel a strong improvement of the system, if it can provide semantically plausible predictions, even though the actual gain in *ksr* might be modest or even slightly decreasing. For this reason we will perform an extended qualitative analysis of the presented methods with persons who use our AAC system *SIBYLLE*. This is one of the main aims of the recently started *ESAC_IMC* project. It is conducted at the *Functional Reeducation and Rehabilitation Centre* of Kerpape, Brittany, where *SIBYLLE* is already used by 20 children suffering from traumatisms of the motor cortex. They appreciate the system not only for communication but also for language learning purposes.

Moreover, we intend to make the word predictor of *SIBYLLE* publicly available (*AFM Voltaire project*) in the not-too-distant future.

## Acknowledgements

This research is partially founded by the UFA (Université Franco-Allemande) and the French foundations APRETREIMC (*ESAC_IMC* project) and AFM (VOLTAIRE project). We also want to thank the developers of the *SRI* and the *Infomap* toolkits for making their programs available.